\documentclass{article}

\usepackage{microtype}
\usepackage{graphicx}
\usepackage{subcaption}
\usepackage{booktabs} 

\usepackage{hyperref}

\usepackage[preprint]{icml2026}

\usepackage{amsmath}
\usepackage{amssymb}
\usepackage{mathtools}
\usepackage{amsthm}
\usepackage{colortbl}  
\usepackage[table]{xcolor} 
\usepackage{multirow}  
\usepackage{wrapfig}
\usepackage{listings}
\usepackage{xcolor}

\usepackage{ifthen}
\usepackage{eso-pic}
\usepackage{graphicx}

\AddToShipoutPictureBG{%
  \ifnum\value{page}=1
    \AtPageUpperLeft{%
      \put(\LenToUnit{2.0cm},\LenToUnit{-2.5cm}){%
        \includegraphics[height=0.8cm]{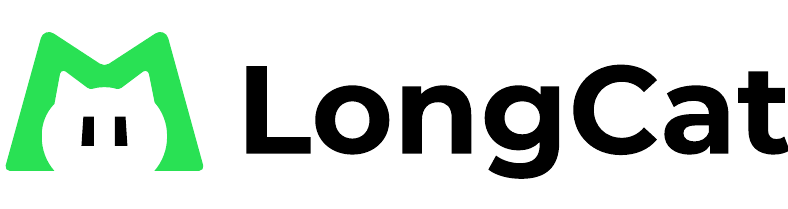}%
      }%
    }%
  \fi
}

\definecolor{softblue}{RGB}{240, 244, 250}
\definecolor{softsage}{RGB}{242, 248, 245}
\definecolor{highlightgold}{RGB}{255, 250, 240}
\usepackage[capitalize,noabbrev]{cleveref}

\theoremstyle{plain}
\newtheorem{theorem}{Theorem}[section]
\newtheorem{proposition}[theorem]{Proposition}

\theoremstyle{definition}
\newtheorem{definition}[theorem]{Definition}

\theoremstyle{remark}

\usepackage[textsize=tiny]{todonotes}

\icmltitlerunning{CoBA-RL: Capability-Oriented Budget Allocation for Reinforcement Learning in LLMs}

\begin{document}

\twocolumn[
  \icmltitle{CoBA-RL: Capability-Oriented Budget Allocation for Reinforcement Learning in LLMs}



  \icmlsetsymbol{equal}{*}
  \icmlsetsymbol{intern}{\dag}

  \begin{icmlauthorlist}
    \newbox\phantombox
    \setbox\phantombox\vbox{\icmlauthor{}{zju,nju,nus,meituan,sjtu}}
    \icmlauthor{Zhiyuan Yao}{zju,meituan,intern}
    \icmlauthor{Yi-Kai Zhang}{nju,meituan}
    \icmlauthor{Yuxin Chen}{nus,meituan}
    \icmlauthor{Yueqing Sun}{meituan}
    \icmlauthor{Zishan Xu}{sjtu}
    \icmlauthor{Yu Yang}{meituan}
    \par
    \icmlauthor{Tianhao Hu}{meituan}
    \icmlauthor{Qi Gu}{meituan}
    \icmlauthor{Hui Su}{meituan}
    \icmlauthor{Xunliang Cai}{meituan}
  \end{icmlauthorlist}

  \icmlaffiliation{zju}{Zhejiang University}
  \icmlaffiliation{nju}{Nanjing University}
  \icmlaffiliation{nus}{National University of Singapore}
  \icmlaffiliation{meituan}{Meituan}
  \icmlaffiliation{sjtu}{Shanghai Jiao Tong University}

  \icmlcorrespondingauthor{Zhiyuan Yao}{yaozhiyuan@zju.edu.cn}
  \icmlcorrespondingauthor{Qi Gu}{guqi03@meituan.com}
  
  \icmlkeywords{Reinforcement Learning, Budget Allocation, Model Capability}

  \vskip 0.3in
  
]




\printAffiliationsAndNotice{\dag Work done during an internship at Meituan.}

\begin{abstract}


Reinforcement Learning with Verifiable Rewards (RLVR) has emerged as a key approach for enhancing LLM reasoning.
However, standard frameworks like Group Relative Policy Optimization (GRPO) typically employ a uniform rollout budget, leading to resource inefficiency. Moreover, existing adaptive methods often rely on instance-level metrics, such as task pass rates, failing to capture the model's dynamic learning state. To address these limitations, we propose CoBA-RL, a reinforcement learning algorithm designed to adaptively allocate rollout budgets based on the model's evolving capability. Specifically, CoBA-RL utilizes a Capability-Oriented Value function to map tasks to their potential training gains and employs a heap-based greedy strategy to efficiently self-calibrate the distribution of computational resources to samples with high training value. Extensive experiments demonstrate that our approach effectively orchestrates the trade-off between exploration and exploitation, delivering consistent generalization improvements across multiple challenging benchmarks. These findings underscore that quantifying sample training value and optimizing budget allocation are pivotal for advancing LLM post-training efficiency. 
Our code is available at \url{https://github.com/Within-yao/CoBA-RL}.

\end{abstract}
\section{Introduction}

\begin{figure}[htbp]
    \centering
    \includegraphics[width=0.98\linewidth]{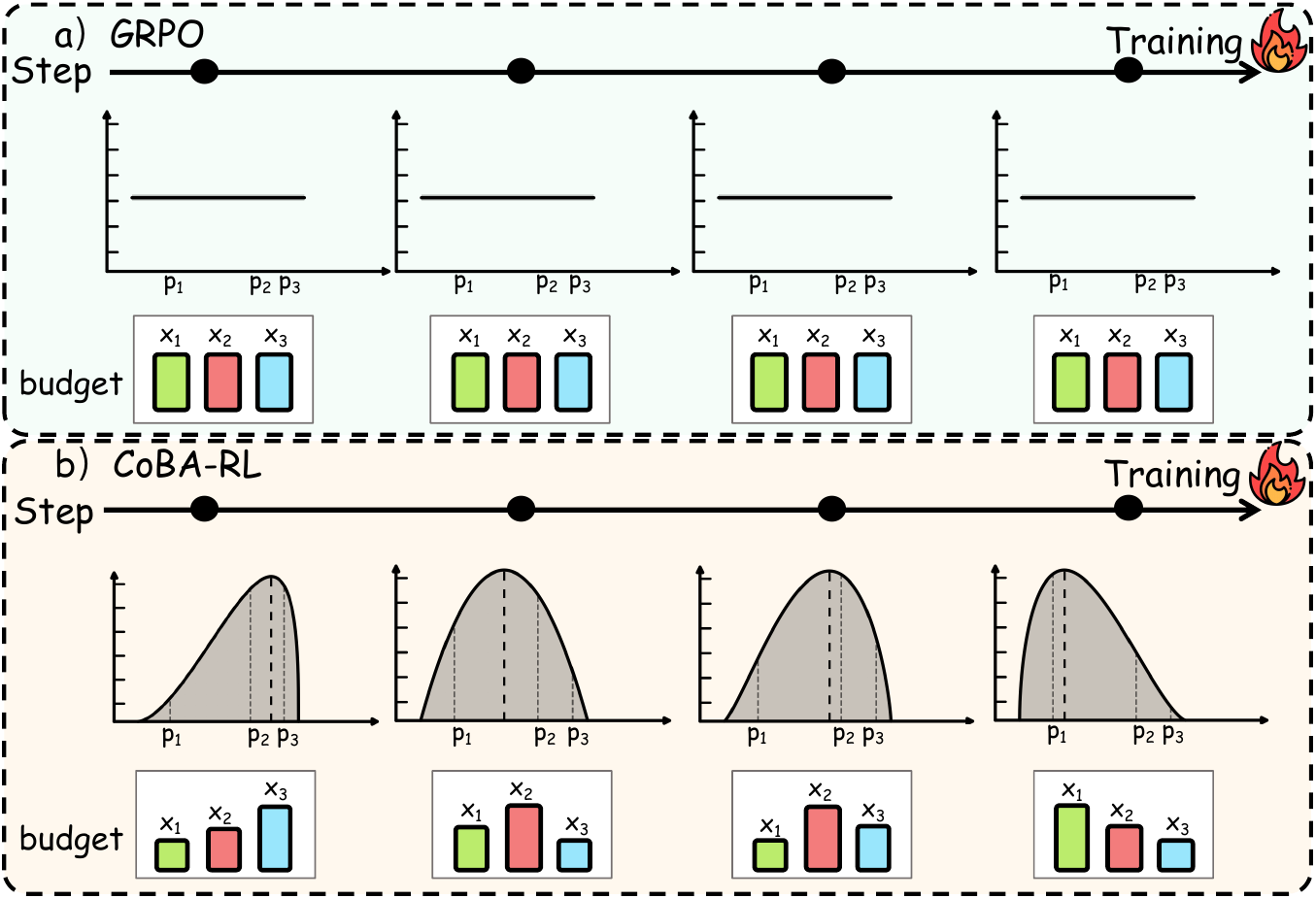}
    \caption{\textbf{Comparison between GRPO-based methods and CoBA-RL.} (a) GRPO employs a uniform strategy independent of training progress. (b) CoBA-RL dynamically self-calibrates the allocation strategy throughout the training process. It autonomously directs the rollout budget toward instances with high training value, aligned with the model's evolving capability. In this visualization, $p_i$ denotes the pass rate corresponding to the task instance $x_i$.}
    \label{fig:moti_case}
\end{figure}

Reinforcement Learning with Verifiable Rewards (RLVR)~\cite{lambert2024tulu} has established itself as a cornerstone for elevating LLM reasoning in agentic, coding, and mathematical domains~\cite{guo2025deepseek,qwen3,comanici2025gemini,kimik2}. Within this landscape, Group Relative Policy Optimization (GRPO)~\cite{shao2024deepseekmath} and its variants~\cite{dapo,gspo,liu2026gdpo,shrivastava2025sample} have risen to prominence. By assigning a uniform budget of $G$ rollouts to every prompt to compute group-relative advantages, GRPO eliminates the need for a separate value network. However, ideally, the rollout budgets allocated to each instance should be commensurate with its training value.Intuitively, complex samples often harbor higher training value and demand extensive exploration, whereas simple instances require minimal resources.

In practice, vanilla GRPO overlooks the critical impact of sample difficulty on training value and the corresponding rollout budget. While recent studies have begun to tailor budgets using instance-level metrics related to task difficulty, such as historical pass rates~\cite{li2025knapsack}, these approaches typically rely on static value functions. They operate on the fixed assumption that harder samples inherently offer superior training value than simpler ones and that this relationship remains constant throughout the entire training process. This perspective neglects a crucial reality: the true training value of a sample is inextricably linked to the policy model's real-time capabilities~\cite{dump,zhang2025preference,wang2025anglesdontlieunlocking}.

As the model’s capabilities evolve during training, the set of samples holding the highest training value constantly shifts~\cite{hu2025vade}. To accommodate these variations, the allocation strategy must continuously calibrate the trade-off between exploitation and exploration. Specifically, exploitation involves consolidating mastery over instances where the model already succeeds, whereas exploration requires allocating resources to sample diverse trajectories on challenging queries, thereby expanding the search space to discover potential solutions~\cite{yang2026your,chen2025pass,hou2025advancing,cui2025entropy}. Consequently, it is imperative to quantify model capability to facilitate policy self-calibration, ensuring the rollout budget is continually re-aligned with the samples most suitable for the current training phase.

To tackle the aforementioned challenges, We propose \textbf{CoBA-RL}, a reinforcement learning algorithm that dynamically allocates the rollout budget in accordance with the model's evolving capability. Specifically, we introduce a Capability-Oriented Value function, modeled as a Beta distribution, to map individual instances to their potential training value. By continuously monitoring the global failure rate of the current training batch, we quantify the model's global capability and dynamically calibrate the shape of the value function, as shown in Figure~\ref{fig:moti_case}. This mechanism autonomously orchestrates the exploration-exploitation trade-off: the value function shifts its high-density regions in real-time to prioritize either consolidating established knowledge or exploring high-uncertainty frontiers based on the model's current competence. Finally, to operationalize this theoretical distribution, we present an Efficient Allocation Optimization algorithm. By formulating the allocation as a constrained maximization problem, we employ a heap-based greedy strategy that iteratively assigns budget to samples offering the highest marginal gain, thereby maximizing the aggregate value of the training batch.


To validate the efficacy of our approach, we conducted extensive experiments using Qwen2.5-7B-Base, Qwen2.5-7B-Instruct, and Qwen3-1.7B/4B-Base models. Empirical results demonstrate that our method significantly outperforms strong baselines across multiple challenging mathematical benchmarks.
These findings suggest that, empowered by the Capability-Oriented Value function, CoBA-RL effectively identifies samples holding the high training value at the current training step, thereby achieving a superior trade-off between exploration and exploitation.

Overall, our contributions are summarized as follows:

\begin{itemize}
    \item We propose a reinforcement learning algorithm that optimizes the exploration-exploitation trade-off by autonomously allocating the rollout budget consistent with the model's evolving capability. By formulating the allocation as a constrained maximization problem, we employ a heap-based greedy strategy to efficiently direct computational resources toward tasks with the highest learning potential.
    
    \item We design a Capability-Oriented Value function as the core allocation criterion. This mechanism dynamically quantifies the training value of each task instance conditioned on the policy model's evolving capability.
    
    \item Extensive experiments across multiple benchmarks validate the effectiveness of \textbf{CoBA-RL}. Our results demonstrate that it significantly outperforms the standard GRPO baseline, as well as static and heuristic strategies, providing a paradigm for efficient Large Language Model post-training.
\end{itemize}



    
    

\section{Method}

\begin{figure*}[!t]
    \centering
    \includegraphics[width=1\textwidth]{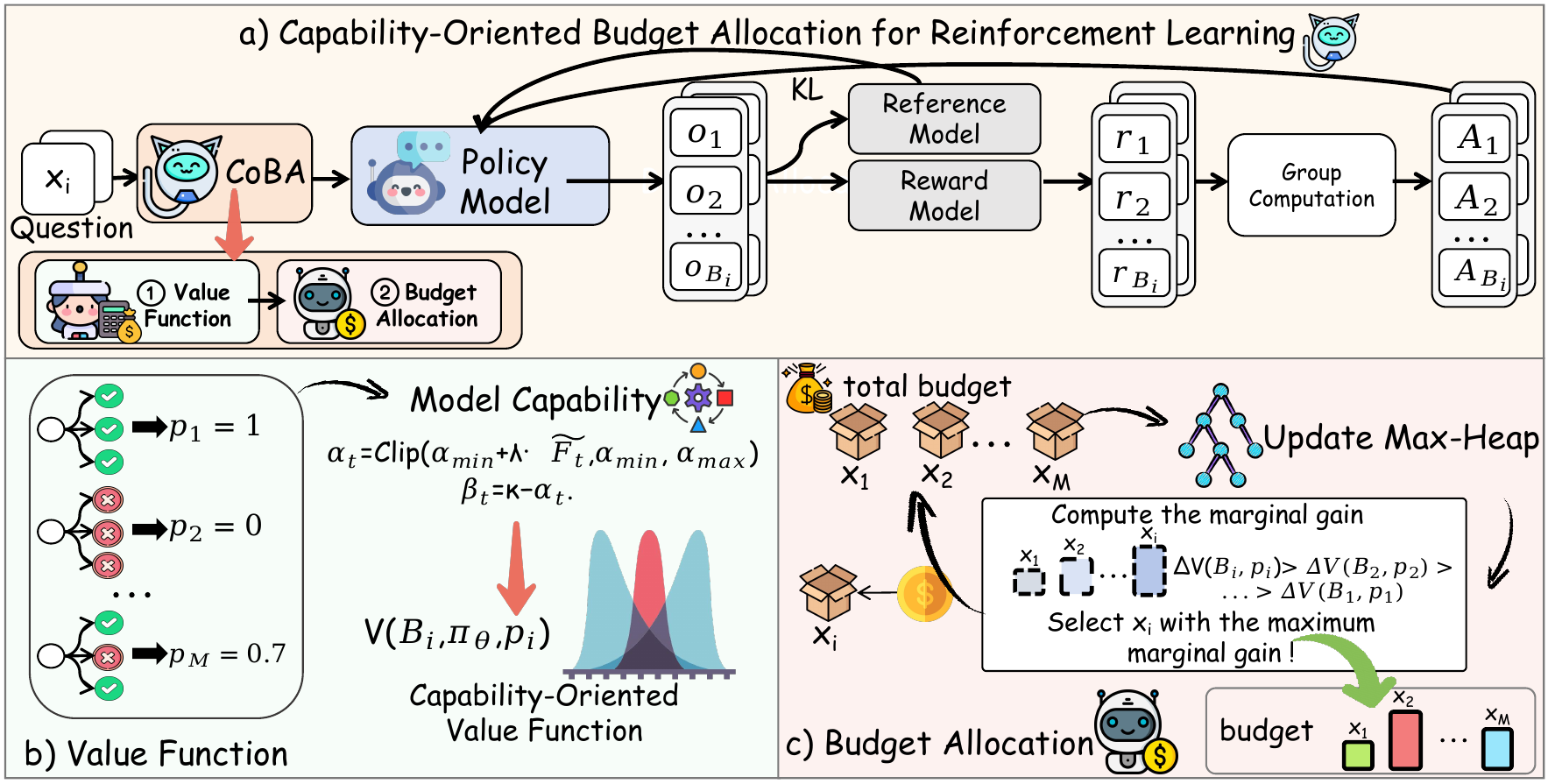}
    \caption{\textbf{Overview of CoBA-RL.} (a) The training pipeline of CoBA-RL. (b) The adaptive value function influenced by model capability. (c) Illustration of the heap-based greedy allocation.}
    \label{fig:framework}
\end{figure*}


In this section, we present CoBA-RL, as illustrated in Figure~\ref{fig:framework}. The framework is composed of two pivotal components. First, the Capability-Oriented Value Function quantifies the policy model's capability via global failure rate statistics. This mechanism self-calibrates the function's shape to discern the model's evolving preferences for specific training samples. Second, the Heap-Based Greedy Budget Allocation is an efficient strategy that optimizes allocation to maximize the cumulative value of the current training batch.



\subsection{Preliminaries}
\label{sec:preliminaries}

The post-training process is formulated as a Reinforcement Learning problem utilizing Group Relative Policy Optimization (GRPO). At each training step $t$, considering a batch of tasks $\mathcal{X}_t = \{x_1, \dots, x_M\}$, the policy $\pi_{\theta_t}$ samples a group of $G$ outputs $\mathcal{O}_i = \{o_{i,1}, \dots, o_{i,G}\}$ for each task $x_i$.

Given a binary outcome reward function $R(x, o) \in \{0, 1\}$, the task pass rate $p_i$ within this group-based generation framework is naturally modeled as the expected probability of correctness:
\begin{equation}
\label{eq:local_pass_rate}
    p_i(x_i; \theta_t) = \mathbb{E}_{o \sim \pi_{\theta_t}(\cdot|x_i)} \left[ \mathbb{I}(R(x_i, o) = 1) \right].
\end{equation}
The optimization objective is to maximize the following loss:
\begin{equation}
\label{eq:grpo_objective}
\begin{aligned}
    \mathcal{J}_{\text{GRPO}}(\theta) &= \mathbb{E}_{x \sim \mathcal{X}_t} \Bigg[ \frac{1}{G} \sum_{k=1}^G \min \Big( \rho_{i,k} A_{i,k}, \\
    & \qquad \qquad \text{clip}(\rho_{i,k}, 1-\epsilon, 1+\epsilon) A_{i,k} \Big) \Bigg],
\end{aligned}
\end{equation}
where $\rho_{i,k}$ denotes the probability ratio between the current and old policies, and $A_{i,k}$ represents the estimated advantage for each output.

\subsection{Problem Formulation and Optimization}
\label{sec:problem_formulation}

We now formally define the budget allocation problem. The system operates under a total rollout budget constraint $B_{\text{total}}$. For each task $x_i$, the policy generates a set of $B_i$ trajectories $\mathcal{O}_i = \{o_{i,1}, \dots, o_{i, B_i}\}$, subject to $\sum B_i = B_{\text{total}}$.

The objective is to determine the optimal set of budgets $\mathbf{B} = \{B_1, \dots, B_M\}$ that maximizes the aggregate training gain defined by $V$. In particular, the function $V$ maps each task $x_i$  and its allocated budget $B_i$ to a scalar value representing the expected learning potential. The optimization problem is formulated as:

\begin{equation}
\label{eq:budget_optimization}
\begin{aligned}
\max_{B_1, \dots, B_M} \quad & \sum_{i=1}^{M} V(B_i, \pi_\theta, p_i) \\
\textrm{subject to} \quad & \sum_{i=1}^{M} B_i = B_{\text{total}}, \\
& B_{\text{low}} \le B_i \le B_{\text{up}}, \quad \forall i \in \{1, \dots, M\}, \\
& B_i \in \mathbb{Z}^{+}.
\end{aligned}
\end{equation}

The primary challenge lies in accurately defining the value function $V$.This function must dynamically reflect the training potential of each task under the current policy $\pi_\theta$. Intuitively, as the model masters an task, its value should diminish, thereby allocating the budget to other samples.

\begin{algorithm}[tb]
   \caption{Heap-Based Greedy Budget Allocation}
   \label{alg:heap_allocation}
   \begin{algorithmic}[1]
   \STATE {\bfseries Input:} Task set $\{x_i\}_{i=1}^M$, Pass rates $p = \{p_i\}$, Total budget $B_{\text{total}}$, Constraints $B_{\text{low}}, B_{\text{up}}$.
   \STATE {\bfseries Output:} Optimal budget allocation $\mathbf{B} = \{B_1, \dots, B_M\}$.
   
   \STATE \textbf{Initialize} $B_i \leftarrow B_{\text{low}}$ for all $i \in \{1, \dots, M\}$.
   \STATE $R \leftarrow B_{\text{total}} - \sum_{i=1}^M B_{\text{low}}$ 
   
   \STATE \textbf{Construct Max-Heap} $\mathcal{H}$:
   \FOR{$i = 1$ {\bfseries to} $M$}
       \IF{$B_i < B_{\text{up}}$}
           \STATE Calculate marginal gain: $\Delta V_i \leftarrow V(B_i + 1, p_i) - V(B_i, p_i)$
           \STATE $\text{Push}(\mathcal{H}, (\Delta V_i, i))$
       \ENDIF
   \ENDFOR
   
   \WHILE{$R > 0$ \AND $\mathcal{H}$ is not empty}
       \STATE $(\Delta V_{i^*}, i^*) \leftarrow \text{Pop}(\mathcal{H})$ \COMMENT{Select task with max marginal gain}
       
       \STATE $B_{i^*} \leftarrow B_{i^*} + 1$
       \IF{$B_{i^*} < B_{\text{up}}$}
           \STATE Update marginal gain: $\Delta V_{\text{new}} \leftarrow V(B_{i^*} + 1, p_{i^*}) - V(B_{i^*}, p_{i^*})$
           \STATE $\text{Push}(\mathcal{H}, (\Delta V_{\text{new}}, i^*))$
       \ENDIF
       
       \STATE $R \leftarrow R - 1$
   \ENDWHILE
   
   \STATE \textbf{return} $\mathbf{B}$
   \end{algorithmic}
\end{algorithm}

\subsection{Capability-Oriented Value Function}
To identify the training samples with high learning value during the training process,, we propose a value function that integrates global capabilities awareness.

\begin{definition}[Global Capability]
\label{def:global_capability}
We formally define the model's global capability at step $t$. The \textbf{Global Success Rate} $\mathcal{S}_t$ is defined as the expected pass rate averaged over the current task batch:
\begin{equation}
    \mathcal{S}_t = \frac{1}{M} \sum_{i=1}^{M} p_i(x_i; \theta_t).
\end{equation}
Correspondingly, the \textbf{Global Failure Rate} $\mathcal{F}_t$ represents the complementary probability: $\mathcal{F}_t = 1 - \mathcal{S}_t$. These metrics serve as quantitative indicators of the policy model's capability variation; specifically, as the model's capability strengthens, the Global Success Rate $\mathcal{S}_t$ is expected to increase.
\end{definition}

\subsubsection{Capability-Induced Preference Density}


To dynamically adjust the model's preference for samples of varying difficulty, we formalize the \textbf{Capability-Induced Preference Density}, which explicitly quantifies the varying degrees of preference that different policy models exhibit toward samples. We model this density as a Beta distribution whose parameters evolve with the training step $t$. We employ the Beta distribution for its flexible parameterization that adaptively shapes the preference density, facilitating a continuous, self-calibrating shift in sampling focus aligned with the model's evolving capability:

\begin{equation}
    \text{Density}(p_i; \alpha_t, \beta_t) = \frac{p_i^{\alpha_t - 1}(1 - p_i)^{\beta_t - 1}}{\mathrm{B}(\alpha_t, \beta_t)},
\end{equation}
where $\alpha_t$ and $\beta_t$ are determined by the current global capability metrics. Specifically, we first compute the moving average of the global failure rate over the past $k$ steps, denoted as $\bar{\mathcal{F}}_t$, to obtain a stable capability estimate. To enhance sensitivity to subtle capability fluctuations during low-failure stages, we apply a non-linear transformation $\Psi(\cdot)$ to obtain $\tilde{\mathcal{F}}_t$:

\begin{equation}
    \tilde{\mathcal{F}}_t = \Psi(\bar{\mathcal{F}}_t) =
    \begin{cases}
    \bar{\mathcal{F}}_t, & \text{if } \bar{\mathcal{F}}_t > 0.5 \\
    \sigma\left(\gamma \cdot (\bar{\mathcal{F}}_t - 0.5)\right), & \text{if } \bar{\mathcal{F}}_t \le 0.5
    \end{cases}
\end{equation}
where $\sigma(\cdot)$ is the Sigmoid function and $\gamma=10$ serves as a scaling factor. Based on $\tilde{\mathcal{F}}_t$, we determine the shape parameters via a linear mapping, while keeping the sum $\kappa = \alpha_t + \beta_t$ constant:

\begin{equation}
\begin{aligned}
    \alpha_t &= \text{clip}\left(\alpha_{\min} + \lambda \cdot \tilde{\mathcal{F}}_t, \ \alpha_{\min}, \ \alpha_{\max}\right), \\
    \beta_t &= \kappa - \alpha_t.
\end{aligned}
\end{equation}

This dynamic modeling mechanism offers significant advantages. Macroscopically, in the early training stages characterized by a high $\tilde{\mathcal{F}}_t$, the distribution skews towards high pass-rate samples to rapidly acquire training signals, gradually shifting towards harder samples as capability improves. Importantly, this mechanism agilely responds to capability fluctuations by automatically recalibrating the preference density, ensuring the budget is allocated to the sample interval that best consolidates the model's current capability, rather than blindly pursuing high-difficulty tasks.

\subsubsection{Budget Saturation Factor}

While the Capability-Induced Preference Density identifies which tasks are theoretically pivotal, the actual training gain derived from a task instance $x_i$ is intrinsically coupled with the allocated computational budget $B_i$. Intuitively, increasing $B_i$ yields higher returns, but this gain follows the law of diminishing returns. To quantify this relationship, we design the \textbf{Budget Saturation Factor}:
\begin{equation}
    \mathcal{\eta}(B_i, p_i) = 1 - e^{-\frac{B_i}{\tau} p_i(1-p_i)},
\end{equation}
where $\tau$ is a temperature coefficient that controls the velocity at which the value reaches saturation. 

By synthesizing the budget constraint with the Capability-Induced Preference Density, we formulate the final \textbf{Capability-Oriented Value Function} as:
\begin{equation}
\label{eq:final_value_function}
    V(B_i, \pi_\theta, p_i) = \left( 1 - e^{-\frac{B_i}{\tau} p_i(1-p_i)} \right) \cdot \text{Density}(p_i; \alpha_t, \beta_t).
\end{equation}


The \textit{Budget Saturation Factor} acts as a realizability coefficient, forcing the optimization of $B_i$ to conform to the shape of the \textit{Capability-Induced Preference Density}. Consequently, as the policy model's capability evolves, thereby altering $\alpha_t$ and $\beta_t$, the objective function $V$ dynamically shifts its topology. This ensures that the budget allocation strategy is not static, but continuously realigns itself with the model's changing proficiency throughout the training trajectory. Through this value function, we can adaptively map different task instances to a value based on the current model capability and the allocated rollout budget.

\subsection{Efficient Allocation Optimization via Heap-Based Greedy Strategy}
\label{method:allocate_method}

\begin{table*}[t]
    \centering
    \caption{Performance comparison of GRPO, Knapsack-RL, and CoBA-RL across different models and benchmarks. All results are reported in percentage (\%). The best results for each model are highlighted in \textbf{bold}. Superscripts denote the difference relative to the GRPO baseline (\textcolor{blue}{Blue} for Knapsack-RL, \textcolor{red}{Red} for CoBA-RL).}
    \label{tab:main_results}
    
    \setlength{\tabcolsep}{12pt} 
    \renewcommand{\arraystretch}{1.3} 
    
    \newcommand{\diffBlue}[1]{\rlap{\ensuremath{^{\textcolor{blue}{\scriptscriptstyle +#1}}}}}
    \newcommand{\diffBlueNeg}[1]{\rlap{\ensuremath{^{\textcolor{blue}{\scriptscriptstyle -#1}}}}}
    \newcommand{\diffRed}[1]{\rlap{\ensuremath{^{\textcolor{red}{\scriptscriptstyle +#1}}}}}
    \newcommand{\diffRedNeg}[1]{\rlap{\ensuremath{^{\textcolor{red}{\scriptscriptstyle -#1}}}}}

    \begin{tabular}{lcccccc<{\hspace{2.5em}}}
        \toprule
        \textbf{Method} & \textbf{AIME24} & \textbf{AIME25} & \textbf{AMC23} & \textbf{MATH500} & \textbf{OLYMPIAD} & \textbf{Avg} \\
        \midrule
        
        \rowcolor{softblue} \multicolumn{7}{c}{\textbf{Qwen2.5-7B-Instruct}} \\
        GRPO & 14.17 & 12.71 & 69.84 & 76.78 & 37.68 & 42.24 \\
        Knapsack-RL & 18.54\diffBlue{4.37} & 15.21\diffBlue{2.50} & 71.41\diffBlue{1.57} & \textbf{80.55}\diffBlue{3.77} & 41.23\diffBlue{3.55} & 45.39\diffBlue{3.15} \\
        \rowcolor{highlightgold}
        CoBA-RL & \textbf{18.96}\diffRed{4.79} & \textbf{18.33}\diffRed{5.62} & \textbf{73.12}\diffRed{3.28} & 80.30\diffRed{3.52} & \textbf{43.19}\diffRed{5.51} & \textbf{46.78}\diffRed{4.54} \\
        
        \midrule

        \rowcolor{softblue} \multicolumn{7}{c}{\textbf{Qwen2.5-7B-Base}} \\
        GRPO & 15.41 & 13.33 & 75.00 & 77.63 & 37.03 & 43.68 \\
        Knapsack-RL & 19.58\diffBlue{4.17} & \textbf{16.67}\diffBlue{3.34} & 74.37\diffBlueNeg{0.63} & 79.73\diffBlue{2.10} & 41.33\diffBlue{4.30} & 46.34\diffBlue{2.66} \\
        \rowcolor{highlightgold}
        CoBA-RL & \textbf{21.04}\diffRed{5.63} & 16.04\diffRed{2.71} & \textbf{76.71}\diffRed{1.71} & \textbf{80.23}\diffRed{2.60} & \textbf{43.11}\diffRed{6.08} & \textbf{47.43}\diffRed{3.75} \\
        
        \midrule
        
        \rowcolor{softblue} \multicolumn{7}{c}{\textbf{Qwen3-4B-Base}} \\
        GRPO & 18.54 & 15.62 & 65.62 & 81.19 & 42.61 & 44.72 \\
        Knapsack-RL & 21.88\diffBlue{3.34} & 20.01\diffBlue{4.39} & 65.47\diffBlueNeg{0.15} & 80.51\diffBlueNeg{0.68} & 45.42\diffBlue{2.81} & 46.66\diffBlue{1.94} \\
        \rowcolor{highlightgold}
        CoBA-RL & \textbf{22.71}\diffRed{4.17} & \textbf{21.16}\diffRed{5.54} & \textbf{72.34}\diffRed{6.72} & \textbf{84.29}\diffRed{3.10} & \textbf{46.78}\diffRed{4.17} & \textbf{49.46}\diffRed{4.74} \\
        
        \midrule
        
        \rowcolor{softblue} \multicolumn{7}{c}{\textbf{Qwen3-1.7B-Base}} \\
        GRPO & 8.96 & 5.00 & 44.69 & 69.46 & 29.92 & 31.61 \\
        Knapsack-RL & 12.50\diffBlue{3.54} & 4.79\diffBlueNeg{0.21} & 45.16\diffBlue{0.47} & 69.69\diffBlue{0.23} & 30.68\diffBlue{0.76} & 32.56\diffBlue{0.95} \\
        \rowcolor{highlightgold}
        CoBA-RL & \textbf{16.25}\diffRed{7.29} & \textbf{7.29}\diffRed{2.29} & \textbf{49.84}\diffRed{5.15} & \textbf{72.85}\diffRed{3.39} & \textbf{32.52}\diffRed{2.60} & \textbf{35.75}\diffRed{4.14} \\
        
        \bottomrule
    \end{tabular}
\end{table*}



The optimization of the objective function defined in Eq.~\ref{eq:budget_optimization} relies on the mathematical property of the proposed value function.


\begin{proposition}
\label{prop:diminishing_returns}
The marginal gain of the value function is strictly monotonically decreasing with respect to the allocated budget $B_i$. That is, defining the marginal gain as $\Delta V(B_i, p_i) = V(B_i +1, p_i) - V(B_i, p_i)$, the following inequality holds for all $B_i \ge 0$:
\begin{equation}
    \Delta V(B_i, p_i) > \Delta V(B_i + 1, p_i).
\end{equation}
\end{proposition}

\begin{figure*}[htbp]
    \centering
    \includegraphics[width=0.99\textwidth]{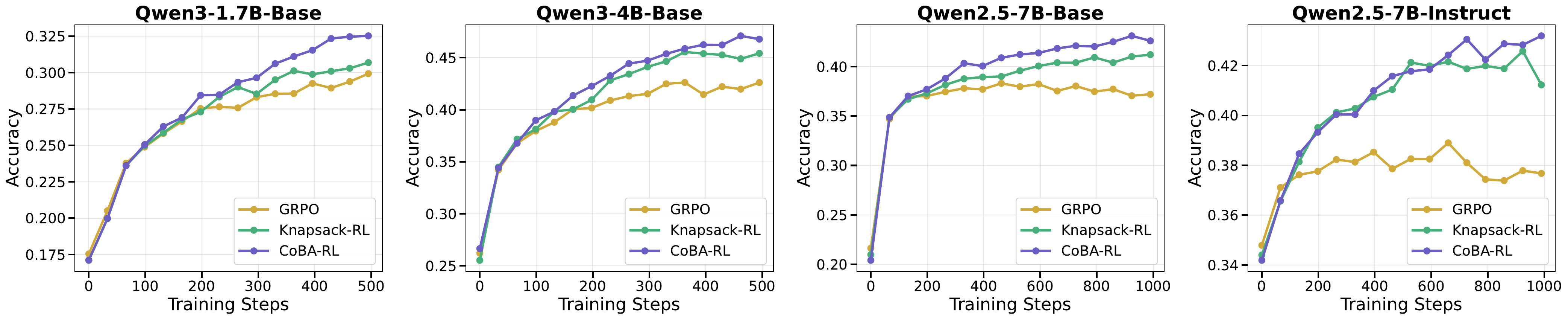}
    \caption{Performance comparison of different models and methods on the Olympiad benchmark (avg@16). The curves track the validation accuracy over training steps for GRPO, Knapsack, and CoBA-RL across varying model scales.}
    \label{fig:olympiad_curves}
\end{figure*}

The proof is provided in the Appendix~\ref{appendix:proof3.2}. This diminishing marginal utility property guarantees that a greedy allocation strategy yields an optimal solution for the discrete resource allocation problem. Consequently, we design an efficient \textbf{Heap-Based Greedy Strategy}. Specifically, we maintain a max-heap of marginal gains for all tasks. In each iteration, we extract the task with the highest potential gain from the heap top, allocate a unit rollout budget, update its state, and re-insert it into the heap. The detailed procedure is outlined in Algorithm~\ref{alg:heap_allocation}.

Notably, this strategy operates based on the relative magnitude of marginal gains within the current training batch. This facilitates a dual-adaptive allocation mechanism where resource distribution naturally shifts in response to the evolving shape of the value function—reflecting the shifting preferences of the current policy $\pi_\theta$—while simultaneously adjusting to the specific variations of the samples in the batch. Such a design guarantees that the computational budget is dynamically channeled toward the instances offering the greatest learning potential.


\section{Experiments}

\subsection{Experimental Setup}
\begin{table}[t]
    \centering
    \caption{Performance comparison of Exploration-Exploitation strategies on Qwen2.5-7B-Instruct.}
    \label{tab:exp_exp_compare}
    \setlength{\tabcolsep}{3.5pt} 
    \renewcommand{\arraystretch}{1.2} 
    
    \resizebox{\columnwidth}{!}{
    \begin{tabular}{lcccccc}
        \toprule
        \textbf{Strategy} & \textbf{AIME24} & \textbf{AIME25} & \textbf{AMC23} & \textbf{MATH500} & \textbf{OLYMPIAD} & \textbf{Avg} \\
        \midrule
        
        Explore $\to$ Exploit & 16.87 & 10.41 & \textbf{73.43} & 79.93 & 41.84 & 44.50 \\
        
        \textbf{Exploit $\to$ Explore (Ours)} & \textbf{18.96} & \textbf{18.33} & 73.12 & \textbf{80.30} & \textbf{43.19} & \textbf{46.78} \\
        
        \bottomrule
    \end{tabular}%
    }
\end{table}

\begin{figure*}[htbp]
    \centering
    \includegraphics[width=0.99\textwidth]{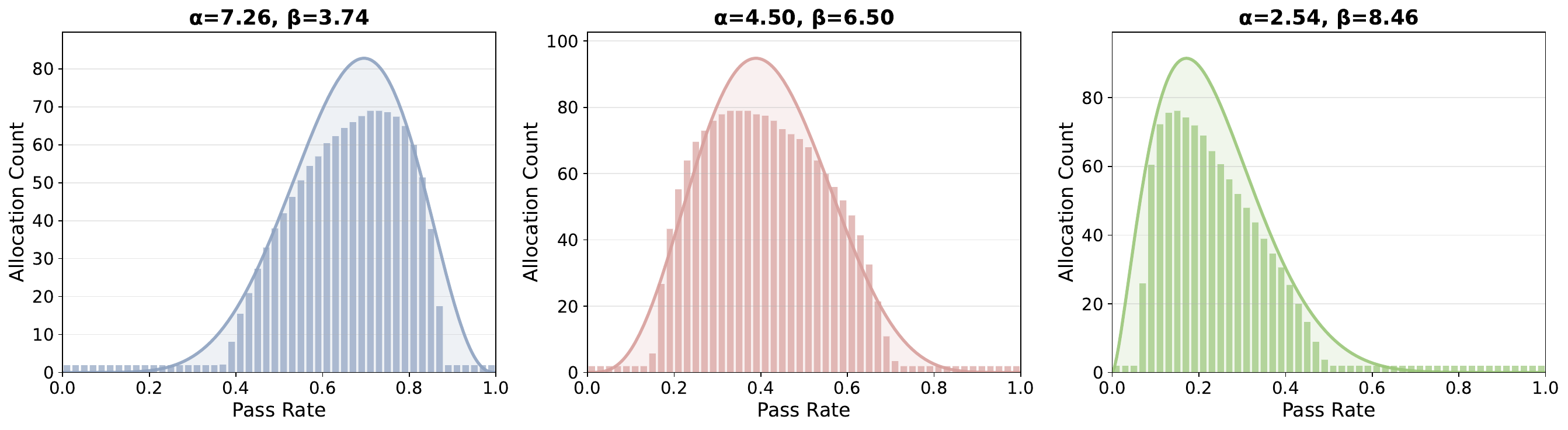}
    \caption{Visualization of different budget allocation across varying model capabilities on Qwen2.5-7B-instruct.}
    \label{fig:beta_allocation_comparison}
\end{figure*}

We initialize our policy models using Qwen2.5-7B-Instruct~\cite{qwen2025qwen25technicalreport}, Qwen2.5-7B-Base~\cite{qwen2025qwen25technicalreport} and Qwen3-1.7B/4B-Base~\cite{qwen3}. For training, we utilize DAPO-Math-17K~\cite{dapo}, a dataset widely adopted for mathematical reasoning tasks. For each training instance, we generate $G=16$ rollout trajectories. 

We evaluate performance on five challenging benchmarks: AIME24~\cite{aime24}, AIME25~\cite{aime25}, AMC23, MATH500~\cite{Math500}, and OLYMPIAD Bench~\cite{OlympiaBench}. To benchmark the efficacy of our dynamic budget allocation, we compare CoBA-RL against two primary baselines: GRPO~\cite{shao2024deepseekmath} and Knapsack-RL~\cite{li2025knapsack}. For fair comparison, all reported results are evaluated using the avg@16 metric. Comprehensive implementation details and additional experimental results are provided in Appendix~\ref{sec:appendix_experiment} and~\ref{appendix:addition}.

\subsection{Main Results}

\begin{table*}[t]
    \centering
    \caption{Quantitative comparison against Static and Heuristic baselines. The best results are highlighted in \textbf{bold}.}
    \label{tab:result_compare_stastic}
    
    \setlength{\tabcolsep}{12pt} 
    \renewcommand{\arraystretch}{1.2} 
    
    \resizebox{0.95\textwidth}{!}{ 
    \begin{tabular}{lcccccc}
        \toprule
        \textbf{Method} & \textbf{AIME24} & \textbf{AIME25} & \textbf{AMC23} & \textbf{MATH500} & \textbf{OLYMPIAD} & \textbf{Avg} \\
        \midrule
        
        $(\alpha=10.5, \beta=1.5)$ & 15.20 & 15.21 & 68.28 & 80.17 & 42.26 & 44.22 \\
        
        $(\alpha=1.5, \beta=10.5)$ & 18.12 & 16.45 & 70.31 & 79.92 & 41.25 & 45.21 \\
        
        Linear Step Decay & 17.08 & 16.85 & 70.77 & 79.81 & 42.43 & 45.39 \\
        \midrule
        
        \rowcolor{highlightgold}
        \textbf{CoBA-RL (Ours)} & \textbf{18.96} & \textbf{18.33} & \textbf{73.12} & \textbf{80.30} & \textbf{43.19} & \textbf{46.78} \\
        
        \bottomrule
    \end{tabular}%
    }
\end{table*}
The quantitative results of our experiments are summarized in Table~\ref{tab:main_results}. In terms of overall performance, CoBA-RL achieves an average accuracy of 46.78\% on Qwen2.5-7B-Instruct. This performance surpasses the GRPO baseline of 42.24\% by a significant margin of 4.54\% and consistently outperforms the 45.39\% accuracy achieved by Knapsack-RL. Parallel improvements are evident on Qwen2.5-7B-Base, where CoBA-RL attains an overall average of 47.43\%. Notably, on the OLYMPIAD benchmark, our method achieves an accuracy of 43.11\%, surpassing the Knapsack-RL baseline of 41.33\% by 1.78\%. Similarly, on the Qwen3-4B-Base and Qwen3-1.7B-Base models, our method yields average improvements of 4.74\% and 4.14\% over GRPO, respectively, demonstrating robust scalability across different model sizes. Figure~\ref{fig:olympiad_curves} illustrates the training curves of CoBA-RL on the Olympiad benchmark (Avg@16) across different models.

Notably, on the AIME25 benchmark using the Qwen2.5-7B-Instruct model, CoBA-RL improves accuracy from 12.71\% to 18.33\%. This marks a 5.62\% increase over GRPO and further surpasses the 15.21\% performance of Knapsack-RL. Furthermore, on the AMC23 dataset with Qwen3-4B-Base, our method achieves a remarkable increase of 6.72\% over the GRPO baseline. These findings underscore the critical importance of dynamically discerning whether the policy requires a bias toward exploration or exploitation throughout the reinforcement learning training process.

To intuitively validate our allocation mechanism, we visualize the resulting budget distributions under different value functions in Figure~\ref{fig:beta_allocation_comparison}.As depicted, the allocated budget dynamically shifts in response to the model's evolving capability, conforming to the geometric shape of the specified value functions.


\subsection{The Relationship between Exploration and Exploitation in Reinforcement Learning}

\begin{figure}[htbp]
    \centering
    \includegraphics[width=1\linewidth]{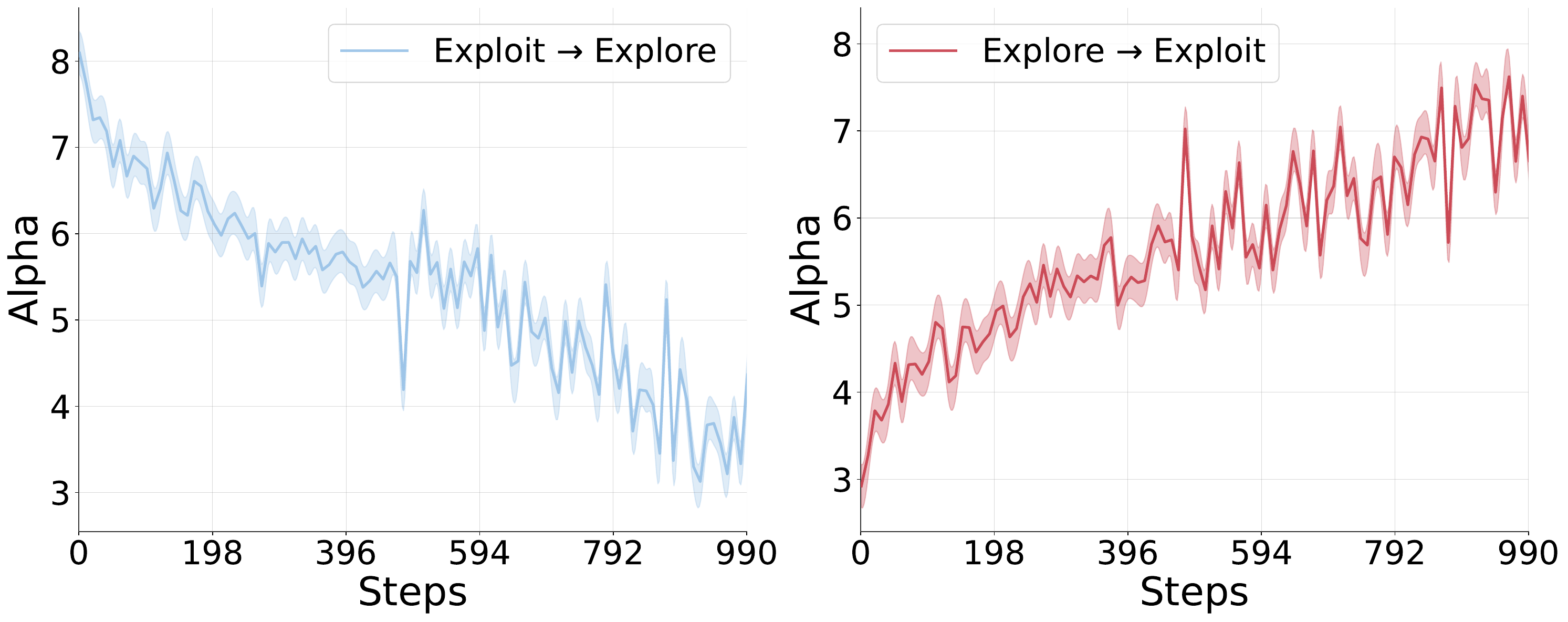}
    \caption{Evolution of $\alpha_t$ on Qwen2.5-7B-Instruct. \textbf{Left:} The "Exploit $\to$ Explore" strategy, where $\alpha_t$ exhibits a fluctuating downward trend. \textbf{Right:} The "Explore $\to$ Exploit" strategy, where $\alpha_t$ shows a fluctuating upward trend.}
    \label{fig:alpha_value}
\end{figure}

We investigate the optimal scheduling of exploration and exploitation within the reinforcement learning process. Specifically, we conducted a comparative experiment on the Qwen2.5-7B-Instruct model to evaluate the performance differences between an ``Explore $\to$ Exploit'' strategy and our proposed ``Exploit $\to$ Explore'' strategy. The evolution of $\alpha_t$ for both methods is visualized in Figure~\ref{fig:alpha_value}.

As presented in Table~\ref{tab:exp_exp_compare}, the results demonstrate that the ``Exploit $\to$ Explore'' strategy yields superior overall performance. Specifically, our method achieves a substantial improvement on the AIME25 benchmark, boosting accuracy from 10.41\% to 18.33\%, and attains the highest average score of 46.78\%. This empirical evidence supports our hypothesis that prioritizing the exploitation of simple samples in the early stages facilitates the rapid consolidation of foundational capabilities, whereas allocating the budget to explore difficult instances in later stages effectively expands the model's solution space.




\subsection{Comparative Analysis against Static and Heuristic Baselines}
\label{exp:compare_static_heuristic}

Our approach explicitly accounts for the evolution of model capability throughout the reinforcement learning process, regulating resource allocation by mapping capability fluctuations to an adaptive value function. To verify the efficacy of this dynamic mechanism, we benchmark it against three baselines categorized into Static and Heuristic strategies. Specifically, we evaluate two Static Strategies employing a Fixed Value Function: one configured with $(\alpha, \beta) = (10.5, 1.5)$ to prioritize exploitation, and another with $(\alpha, \beta) = (1.5, 10.5)$ to prioritize exploration. Additionally, we compare against a Heuristic Strategy using Linear Step Decay, where $\alpha_t$ decreases stepwise from 10 to 1 (i.e., $10 \to 9 \to \dots \to 1$) corresponding to the progression of training steps.Visualizations of these strategies are provided in Appendix~\ref{app:baseline_viz}.

The quantitative results are presented in Table~\ref{tab:result_compare_stastic}. As observed, our dynamic budget allocation strategy consistently outperforms both the static fixed-parameter strategies and the heuristic baseline. CoBA-RL achieves the highest average accuracy of 46.78\%, surpassing the 45.39\% performance of Linear Step Decay and the 45.21\% accuracy of the best Static baseline by margins of 1.39\% and 1.57\%, respectively. 
This validates that static, pre-determined strategies struggle to align with the unpredictable dynamics of real learning compared to our adaptive approach.




\subsection{Ablation Study}

\begin{figure}[htbp]
    \centering
    \includegraphics[width=1\linewidth]{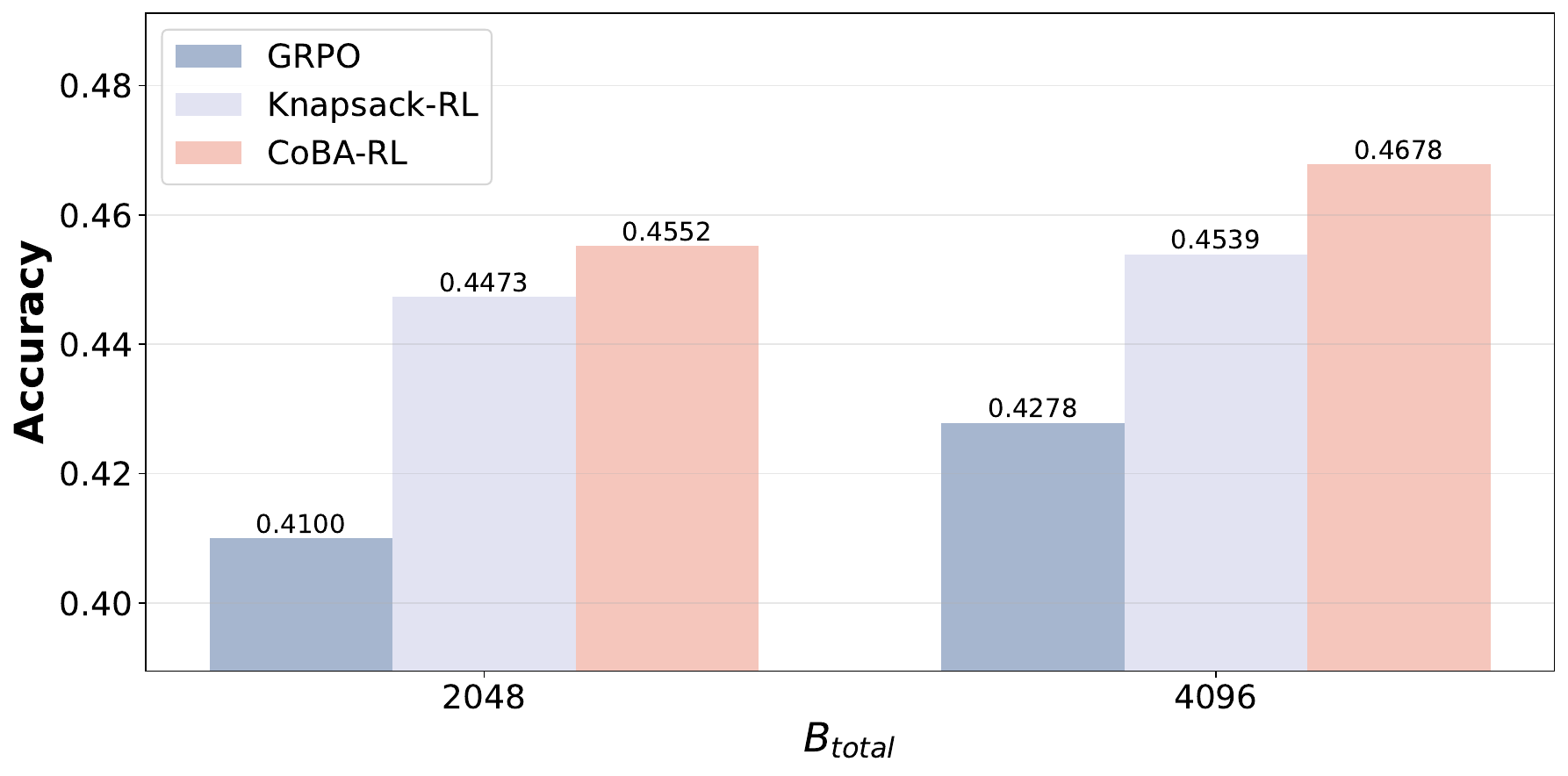}
    \caption{Performance comparison under different exploration budget.}
    \label{fig:group_ablation}
\end{figure}

We evaluate the performance of Qwen2.5-7B-instruct under varying total budget constraints $B_{\text{total}}$, as illustrated in Figure~\ref{fig:group_ablation}. The results clearly demonstrate the consistent superiority of CoBA-RL over both the baseline GRPO and Knapsack-RL across different resource constraints. Significantly, CoBA-RL exhibits exceptional data efficiency; with a restricted budget of $B_{\text{total}}=2048$, it attains an accuracy of 45.52\%. This result not only exceeds the performance of competing methods at the same budget level but, strikingly, surpasses the 42.78\% accuracy of GRPO trained with a doubled budget of $B_{\text{total}}=4096$. We attribute this to the inherent resource inefficiency of GRPO's uniform allocation strategy, which indiscriminately distributes the doubled budget, thereby squandering significant computational resources on samples with negligible training potential.In contrast, our findings verify that the capability-oriented value function in CoBA-RL effectively guides the model to utilize computational resources to explore high-value instances, thereby maximizing the aggregate value of the training batch.





\subsection{Runtime Efficiency of Budget Allocation}
\begin{table}[h]
\centering
\caption{Computational efficiency comparison between DP and our Heap-Based Greedy strategy ($B_{\text{total}}=8192$).}
\label{tab:efficiency}
\resizebox{1.0\columnwidth}{!}{
\begin{tabular}{l|ccc}
\toprule
 & \textbf{DP} & \textbf{Heap-Based Greedy (Ours)} & \textbf{Speedup} \\
\midrule
\textbf{Allocation Time (s)} & 115.05 & \textbf{0.124} & $\sim 928\times$ \\
\bottomrule
\end{tabular}
}
\end{table}



To validate the computational superiority of our allocation algorithm, we compare the Heap-Based Greedy strategy against the standard Dynamic Programming baseline in Table~\ref{tab:efficiency}. Notably, due to the diminishing marginal returns property proven in Appendix~\ref{appendix:proof3.2}, both methods theoretically yield identical allocation results; thus, our comparison focuses exclusively on runtime.

Theoretical analysis reveals that the Dynamic Programming baseline suffers from a pseudo-polynomial complexity of $O(M \cdot B_{\text{total}} \cdot (B_{\text{up}} - B_{\text{low}}))$. Consequently, the computational cost scales linearly with the product of the batch size, the total budget, and the task-specific allocation range. Given a large state space where the batch size $M$ is 512 and the total budget $B_{\text{total}}$ equals 8192, the baseline method requires approximately 115.05 seconds. This latency is unacceptable for online reinforcement learning loops. In contrast, our Heap-Based Greedy strategy significantly reduces the complexity to $O(B_{\text{total}} \log M)$, effectively decoupling the multiplicative dependency between budget size and batch size. Consequently, our method completes the allocation in merely 0.124 seconds. This demonstrates that our approach operates with minimal time complexity, allowing for seamless integration into large-scale training pipelines.


\section{Related Work}

\subsection{Reinforcement Learning for LLMs}
Methods based on Reinforcement Learning with Verification and Reasoning (RLVR) have proven effective in enhancing the reasoning capabilities of LLMs~\cite{dai2025cde,trung2024reft,zheng2025parallel,zheng2025learning,jaech2024openai,xie2025logic}. Among them, GRPO~\cite{shao2024deepseekmath} has been widely adopted due to its effectiveness and efficiency. Building on this, GSPO~\cite{gspo} defines sample importance based on sequence likelihood, while DAPO~\cite{dapo} introduces four distinct techniques to bolster reinforcement learning performance. 
However, these predominantly group-based mechanisms often overlook the inherent variability across different tasks, inevitably leading to a significant waste of rollout resources.

\subsection{Progressive Training and Resource Allocation}

Progressive training strategies, particularly curriculum learning, have been widely adopted to enhance model performance by organizing training data into distinct difficulty stages~\cite{shi2025efficient,wu2025progressive,zeng2025cures}. These approaches typically adhere to an easy-to-hard paradigm, emphasizing sample selection and curriculum design~\cite{lee2024instruction,nair2024curriculum,deng2025boosting,li2025adacurl}. For instance, ADCL~\cite{zhang2025learning} addresses difficulty shifts by periodically evaluating subsequent data batches, while SEC~\cite{chen2025self} utilizes policy gradient advantages to dynamically adjust data distribution. While curriculum methods focus on selecting which samples to train, our approach focuses on adaptively allocating varying budgets.

To achieve this, we draw inspiration from budget allocation, a fundamental problem in operations research~\cite{hussain2013survey}. Extensive studies have explored this in domains such as online advertising and marketing~\cite{liu2020effective,chen2025multichannel,marketAllocation}, often utilizing Multi-Armed Bandit frameworks to optimize resource distribution~\cite{ge2025multi}. Recently, these concepts have been adapted to Large Language Models (LLMs). For example, ROI-Reasoning~\cite{zhao2026roi} formulates inference under limited token budgets as an ordered stochastic multiple-choice knapsack problem.

Despite these advancements, dynamically allocating rollout budgets during reinforcement learning (RL)—specifically to optimize the trade-off between exploration and exploitation—remains a significant challenge. While approaches like GVM-RAFT~\cite{yao2025optimizing} allocate resources via rejection sampling to minimize stochastic gradient variance, and Knapsack-RL~\cite{li2025knapsack} employs a classic knapsack formulation~\cite{pisinger1998knapsack} to maximize batch value, they often rely on static or pre-defined value functions. Consequently, they fail to effectively adapt to the model's capabilities, which evolve dynamically throughout the training process, as they lack a mechanism to explicitly correlate the potential training value of individual samples with the model's real-time proficiency.
\section{Conclusion}



We propose CoBA-RL, a reinforcement learning algorithm that dynamically adapts rollout budget allocation to evolving LLM capabilities by formulating the task as a constrained optimization problem. Importantly, this algorithm allows for seamless integration into existing RL pipelines. To precisely quantify the learning value of different samples during training, we introduce a Capability-Oriented Value function modeled via a dynamic Beta distribution, employing a heap-based greedy strategy to iteratively maximize marginal gains. Experiments demonstrate that CoBA-RL achieves a superior exploration-exploitation trade-off, significantly outperforming static and heuristic baselines. 
Our analysis further reveals that effective training dynamics typically transition from exploitation to exploration, shifting from the rapid improvement of model capabilities to the investigation of diverse trajectories within a larger search space. Looking forward, we posit that accurately defining the training potential of tasks and optimizing budget allocation represent critical directions for the future advancement of efficient LLM post-training.




\bibliography{main}
\bibliographystyle{icml2026}

\newpage
\appendix
\onecolumn
\section{Proof of Proposition 3.2}
\label{appendix:proof3.2}

Recall the definition of the value function given in Eq. \ref{eq:final_value_function}:
\begin{equation}
    V(B_i, \pi_\theta, p_i) = \left( 1 - e^{-\frac{B_i}{\tau} p_i(1-p_i)} \right) \cdot \text{Density}(p_i; \alpha_t, \beta_t).
\end{equation}
For simplicity, let us denote the terms independent of $B_i$ as constants $C$ and $k$. Let:
\begin{align}
    C &= \text{Density}(p_i; \alpha_t, \beta_t) = \frac{p_i^{\alpha_t - 1}(1 - p_i)^{\beta_t - 1}}{\mathrm{B}(\alpha_t, \beta_t)}, \\
    k &= \frac{p_i(1-p_i)}{\tau}.
\end{align}
We consider the non-trivial case where $p_i \in (0, 1)$ and $\tau > 0$, implying that $C > 0$ and $k > 0$. The value function can be rewritten as:
\begin{equation}
    V(B_i, \pi_\theta, p_i) = C \left( 1 - e^{-k B_i} \right).
\end{equation}

\paragraph{Step 1: Deriving the Marginal Gain.}
The marginal gain $\Delta V(B_i, p_i)$ represents the increment in value obtained by increasing the budget by one unit:
\begin{align}
    \Delta V(B_i, p_i) &= V(B_i + 1, \pi_\theta, p_i) - V(B_i, \pi_\theta, p_i) \nonumber \\
    &= C \left( 1 - e^{-k (B_i + 1)} \right) - C \left( 1 - e^{-k B_i} \right) \nonumber \\
    &= C \left( e^{-k B_i} - e^{-k (B_i + 1)} \right) \nonumber \\
    &= C e^{-k B_i} \left( 1 - e^{-k} \right).
\end{align}
Let $A = C(1 - e^{-k})$. Since $k > 0$, we have $e^{-k} < 1$, thus $(1 - e^{-k}) > 0$. Combined with $C > 0$, it follows that $A > 0$. The marginal gain simplifies to:
\begin{equation}
    \Delta V(B_i, p_i) = A \cdot e^{-k B_i}.
\end{equation}

\paragraph{Step 2: Proving Monotonicity.}
To prove that the marginal gain is strictly decreasing, we compare $\Delta V(B_i, p_i)$ with $\Delta V(B_i + 1, p_i)$. We examine the ratio between consecutive marginal gains:
\begin{equation}
    \frac{\Delta V(B_i + 1, p_i)}{\Delta V(B_i, p_i)} = \frac{A \cdot e^{-k (B_i + 1)}}{A \cdot e^{-k B_i}} = \frac{e^{-k} \cdot e^{-k B_i}}{e^{-k B_i}} = e^{-k}.
\end{equation}
Since $k > 0$, we have:
\begin{equation}
    e^{-k} < 1.
\end{equation}
Therefore:
\begin{equation}
    \frac{\Delta V(B_i + 1, p_i)}{\Delta V(B_i, p_i)} < 1 \quad \implies \quad \Delta V(B_i + 1, p_i) < \Delta V(B_i, p_i).
\end{equation}
This confirms that the marginal gain $\Delta V(B_i, p_i)$ is a strictly decreasing geometric sequence with respect to the budget $B_i$.

\section{Implementation Details of Main Training Loop}
\label{app:code_implementation}

Listing~\ref{lst:allocator_implementation} illustrates how the CoBA-RL BudgetAllocator is integrated into the GRPO training loop. The core logic involves calculating the specific rollout count for each prompt in the current batch via a dictionary mapping and resampling the batch accordingly before the generation phase.

\begin{lstlisting}[caption={Core logic for Budget Allocation integration in GRPO.}, label={lst:allocator_implementation}]
# Inside the GRPO training loop
for batch_dict in dataloader:
    # Extract Batch
    batch = process_batch(batch_dict)
    indices = batch["index"]

    # Execute Budget Allocation
    # Returns a dictionary: {index: count}
    total_budget = batch_size * default_group_size
    allocation_dict = budget_allocator.allocate(
        indices=indices, 
        total_budget=total_budget
    )

    # Expand batch according to allocation counts
    repeat_indices = []
    for idx, count in allocation_dict.items():
        if count > 0:
            repeat_indices.extend([idx] * count)

    batch = batch[repeat_indices]
    
    outputs = actor.generate_sequences(batch)
    # ... GRPO loss calculation and update ...
\end{lstlisting}

\section{Experiment Details}
\label{sec:appendix_experiment}

We implement our training pipeline using the \textbf{Verl}~\cite{sheng2024hybridflow} framework, utilizing SGLang~\cite{zheng2024sglang} as the inference engine. For optimization, we employ the AdamW optimizer with a learning rate of $1 \times 10^{-6}$. The hyperparameters are adjusted based on model scale: for models smaller than 7B, we set the global batch size $M=512$ and train for approximately 500 steps; for the 7B model, we set $M=256$ and extend training to nearly 1000 steps. To accommodate complex reasoning chains, we set the maximum response length to 4096 tokens. Consistent with recent trends in reasoning alignment, we do not incorporate the KL divergence penalty (i.e., $\beta_{\text{KL}} = 0$). Regarding the specific parameters for our \textbf{CoBA-RL} method, we constrain the rollout budget for each instance within the range $[B_{\text{low}}, B_{\text{up}}] = [2, 128]$. The sensitivity analysis of the parameter $\kappa$ is detailed in Appendix~\ref{app:sen_kappa}. The detailed inference hyperparameters used for evaluation are listed in Table~\ref{tab:eval_params}. Notably, since the official implementation of the baseline Knapsack-RL is not open-sourced, we re-implement it within the Verl.

\begin{table}[htbp]
    \centering
    \caption{Hyperparameters for evaluation rollout.}
    \label{tab:eval_params}
    \begin{tabular}{lc}
        \toprule
        \textbf{Hyperparameter} & \textbf{Value} \\
        \midrule
        Temperature & 1.0 \\
        Top-p & 0.9 \\
        Sample Count ($n$) & 16 \\
        Do Sample & True \\
        \bottomrule
    \end{tabular}
\end{table}

Regarding the specific implementation of budget allocation methods, following the recommendations of DAPO~\cite{dapo}, we adopt the \textit{Clip-higher} strategy for both Knapsack-RL and CoBA-RL.

\section{Additional Results}
\label{appendix:addition}

\subsection{Analysis of Task Difficulty Transition}

\begin{figure*}[!t]
    \centering
    \includegraphics[width=0.99\textwidth]{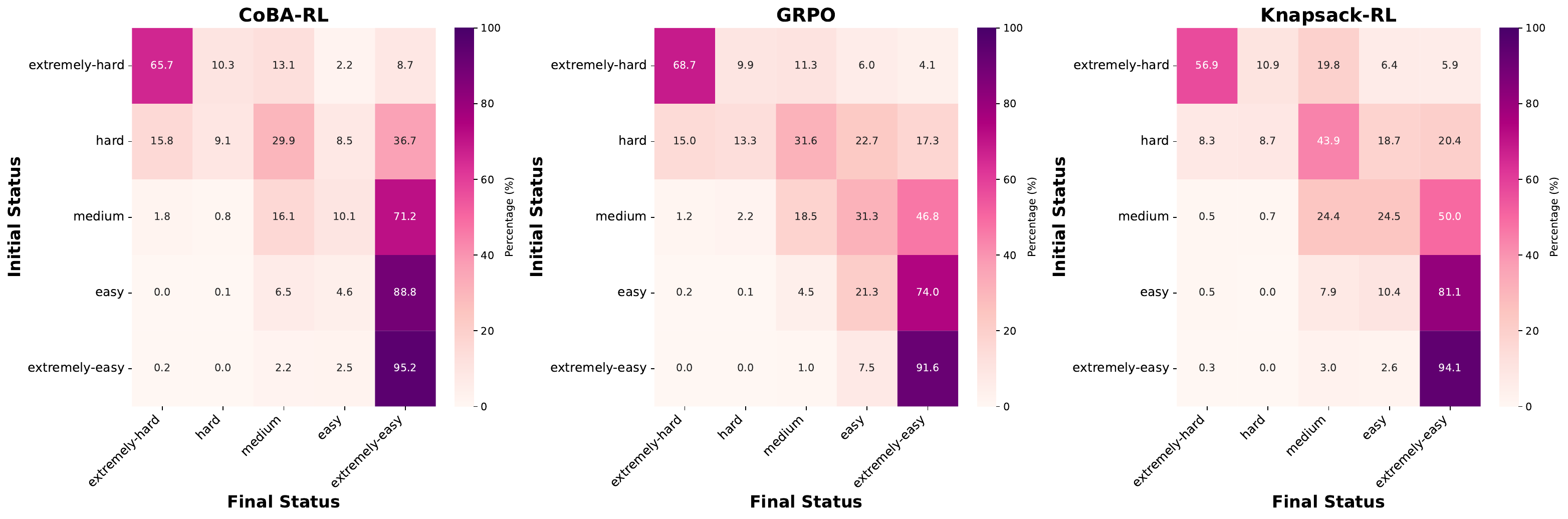}
    \caption{Task transition matrices for Qwen2.5-7B-Instruct during training. The cell $(i, j)$ indicates the percentage of samples transitioning from the initial status $i$ to the final status $j$. }
    \label{fig:task_transition}
\end{figure*}

To evaluate performance across different learning stages, we categorize training prompts into five difficulty levels based on initial success rate ($p_i$): \textit{extremely-hard} ($p_i=0$), \textit{hard} ($0 < p_i \le 0.2$), \textit{medium} ($0.2 < p_i < 0.8$), \textit{easy} ($0.8 \le p_i < 1.0$), and \textit{extremely-easy} ($p_i=1.0$). As shown in the final column of Figure~\ref{fig:task_transition}, CoBA-RL consistently achieves the highest conversion rates. For \textit{medium} tasks, it reaches 71.2\%, significantly surpassing GRPO (46.8\%) and Knapsack-RL (50.0\%). For \textit{hard} tasks, it achieves 36.7\%, nearly doubling the GRPO baseline (17.3\%) and outperforming Knapsack-RL (20.4\%). Furthermore, it leads in converting \textit{extremely-hard} tasks (8.7\% vs. 4.1\%) and \textit{easy} tasks (88.8\% vs. 74.0\%), while maintaining the highest retention for \textit{extremely-easy} instances (95.2\%).

These results underscore the distinct superiority of CoBA-RL:  By dynamically aligning resources with model capability, CoBA-RL maximizes learning potential across the entire difficulty spectrum, from solving novel problems to retaining 
established knowledge.

\subsection{Sensitivity Analysis of $\kappa$}
\label{app:sen_kappa}

\begin{figure}[htbp]
    \centering
    \includegraphics[width=0.75\linewidth]{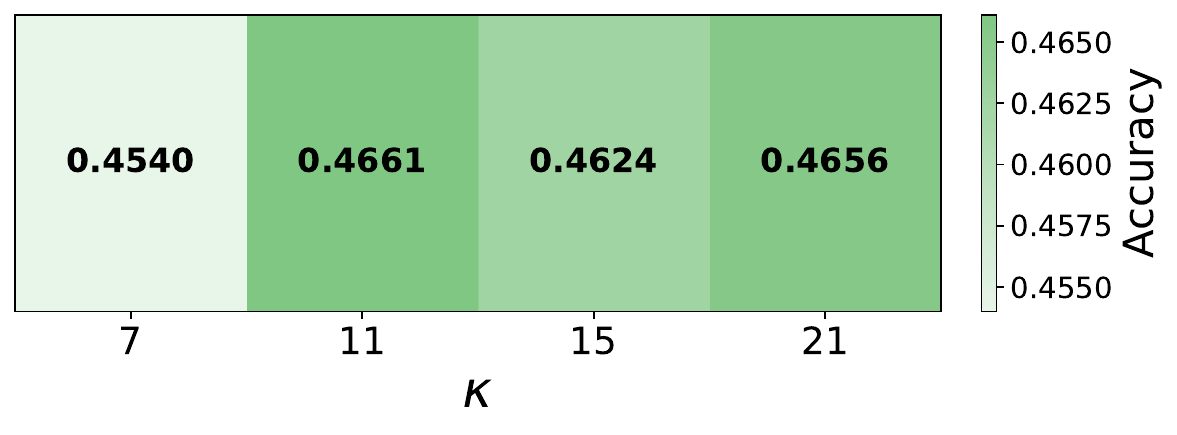}
    \caption{Ablation study results illustrating the impact of different sum parameter $\kappa \in \{7, 11, 15,21\}$ on model performance.}
    \label{ablation:beta_sum}
\end{figure}


In this section, we examine the impact of the hyperparameter $\kappa$---defined as the constant sum of the Beta distribution shape parameters ($\alpha + \beta$)---on the model training process. This parameter controls the variance of the Beta distribution, thereby influencing the "sharpness" of the capability-oriented value function.

We evaluate the performance across four distinct values: $\kappa \in \{7, 11, 15, 21\}$. As illustrated in Figure~\ref{ablation:beta_sum}, our method exhibits strong robustness to variations in $\kappa$, with performance fluctuations remaining minimal across different settings. Specifically, the accuracy ranges from $45.40\%$ (at $\kappa=7$) to a peak of $46.61\%$ (at $\kappa=11$).

This stability validates the reliability of our capability-driven allocation criterion, suggesting that the improvement stems from the dynamic mechanism itself rather than specific hyperparameter tuning. Based on these results, we adopt $\kappa=11$ as the default configuration for our main experiments to achieve optimal performance.

\subsection{Visualization of Static and Heuristic Baselines}
\label{app:baseline_viz}

\begin{figure}[htbp]
    \centering
    \includegraphics[width=0.99\linewidth]{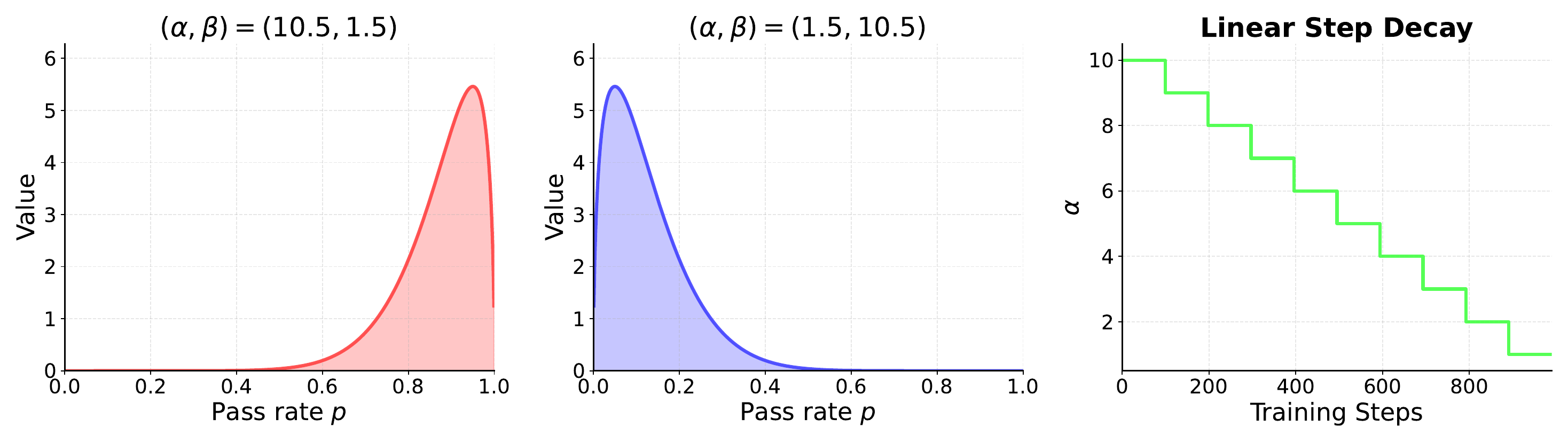}
    \caption{Visualization of the baseline strategies. \textbf{Left \& Middle:} The probability density functions of the static Beta distributions used for exploitation ($(\alpha, \beta)=(10.5, 1.5)$) and exploration ($(\alpha, \beta)=(1.5, 10.5)$). \textbf{Right:} The pre-defined annealing schedule for $\alpha$ in the Linear Step Decay heuristic baseline.}
    \label{fig:baseline_strategies}
\end{figure}

To facilitate a deeper understanding of the comparative analysis presented in the section~\ref{exp:compare_static_heuristic}, we provide visual illustrations of the baseline strategies in Figure~\ref{fig:baseline_strategies}. 

Figure~\ref{fig:baseline_strategies} (Left and Middle) depicts the geometric shapes of the fixed value functions used in the Static Strategies. The configuration $(\alpha, \beta) = (10.5, 1.5)$ results in a distribution heavily skewed towards high success rates $p$, thereby enforcing an exploitation-prioritized allocation. Conversely, the setting $(\alpha, \beta) = (1.5, 10.5)$ yields a distribution peaked at low success rates, promoting an exploration-prioritized strategy. 

Figure~\ref{fig:baseline_strategies} (Right) illustrates the Heuristic Strategy(Linear Step Decay), where the parameter $\alpha_t$ is annealed stepwise from 10 to 1 over the course of training. This heuristic implements a linear, stepwise decrease in $\alpha$, yet it follows a rigid, pre-defined schedule that lacks the adaptability of our capability-oriented approach.

\end{document}